\documentclass[11pt]{article}

\PassOptionsToPackage{numbers, compress}{natbib}

\usepackage{natbib}

\usepackage{hyperref}       %

\hypersetup{
    colorlinks=true,
    linkcolor=blue,
    urlcolor=magenta,
    citecolor=blue}

\date{} %
\advance \oddsidemargin by -0.8in
\newcommand{\myparskip}{3pt}
\parskip \myparskip

\newif\ifcomments
\commentstrue
\ifcomments
    \providecommand{\sameer}[2][]{{\protect\color{violet}{[\textbf{sameer}:\textbf{#1} #2]}}}
    \providecommand{\dylan}[2][]{{\protect\color{purple}{[\textbf{dylan}:\textbf{#1} #2]}}}
    \providecommand{\hima}[2][]{{\protect\color{blue}{[\textbf{hima}:\textbf{#1} #2]}}}
    \providecommand{\yuxin}[2][]{{\protect\color{brown}{[\textbf{yuxin}:\textbf{#1} #2]}}}
    \providecommand{\chenhao}[2][]{{\protect\color{magenta}{[\textbf{chenhao}:\textbf{#1} #2]}}}
\else
    \providecommand{\sameer}[2][]{}
    \providecommand{\dylan}[2][]{}
    \providecommand{\hima}[2][]{}
    \providecommand{\yuxin}[2][]{}
    \providecommand{\chenhao}[2][]{}
\fi

\usepackage[utf8]{inputenc} %
\usepackage[T1]{fontenc}    %
\usepackage{hyperref}       %
\usepackage{url}            %
\usepackage{booktabs}       %
\usepackage{amsfonts}       %
\usepackage{nicefrac}       %
\usepackage{microtype}      %
\usepackage[usenames,dvipsnames,svgnames,x11names]{xcolor}
\usepackage{selectp}
\usepackage{diagbox}
\usepackage{graphicx}
\usepackage{subcaption}
\usepackage{amsmath,amsthm,amsfonts}
\usepackage{textcomp}
\usepackage{amssymb}
\usepackage{algpseudocode,algorithm,algorithmicx}
\usepackage{amsmath}
\usepackage{soul}
\usepackage{sidecap}
\usepackage{enumitem} 
\AtBeginDocument{%
  \providecommand\BibTeX{{%
    \normalfont B\kern-0.5em{\scshape i\kern-0.25em b}\kern-0.8em\TeX}}}

\usepackage{graphicx}
\title{Rethinking Explainability as a Dialogue: A Practitioner's Perspective}
\author{%
  Himabindu Lakkaraju\thanks{Equal Contribution} \\
  Harvard University \\
  \texttt{hlakkaraju@hbs.edu} \\
  \and
  Dylan Slack\footnotemark[1] \\
  UC Irvine \\
  \texttt{dslack@uci.edu} \\
  \and
  Yuxin Chen \\
  University of Chicago \\
  \texttt{chenyuxin@uchicago.edu} \\
  \and
  Chenhao Tan \\
  University of Chicago \\
  \texttt{chenhao@uchicago.edu} \\
  \and
  Sameer Singh \\
  UC Irvine / AI2 \\
  \texttt{sameer@uci.edu} \\
 }
\begin{document}
\maketitle

\begin{abstract}
As practitioners increasingly deploy machine learning models in critical domains such as healthcare, finance, and policy, it becomes vital to ensure that domain experts function effectively alongside these models.
Explainability is one way to bridge the gap between human decision-makers and machine learning models.
However, most of the existing work on explainability focuses on one-off, static explanations like feature importances or rule-lists.
These sorts of explanations may not be sufficient for many use cases that require dynamic, continuous discovery from stakeholders that have a range of skills and expertise.
  In the literature, few works ask decision-makers such as doctors, healthcare professionals, and policymakers about the utility of existing explanations and other desiderata they would like to see in an explanation going forward. 
  In this work, we address this gap and carry out a study where we interview doctors, healthcare professionals, and policymakers about their needs and desires for explanations. Our study indicates that decision-makers would strongly prefer \textit{interactive} explanations.
  In particular, they would prefer these interactions to take the form of \textit{natural language dialogues}.
  Domain experts wish to treat machine learning models as ``another colleague'', i.e., one who can be held accountable by asking \textit{why} they made a particular decision through expressive and accessible natural language interactions.
  Considering these needs, we outline a set of five principles researchers should follow when designing interactive explanations as a starting place for future work.
  Further, we show why natural language dialogues satisfy these principles and are a desirable way to build interactive explanations.
  Next, we provide a design of a dialogue system for explainability, and discuss the risks, trade-offs, and research opportunities of building these systems.
Overall, we hope our work serves as a starting place for researchers and engineers to design interactive, natural language dialogue systems for explainability that better serve users' needs.
  
\end{abstract}

\maketitle

\section{Introduction}

As engineers, researchers, and domain experts increasingly deploy machine learning models in societally critical domains, such as healthcare, criminal justice, and public policy, there is an ever-growing demand for explainability of these models \cite{lipton2018mythos,doshi2017towards,kim2017interpretability,tan2018distill,guidotti2018survey}.
Researchers have proposed a variety of approaches to address this demand for explainability.
For example, a popular class of approaches identifies feature importance, i.e., how much each feature contributes to the model prediction \citep{ribeiro2016should,lundberg2017unified}.
The seminal work, LIME, shows that such feature importance can help people understand why a model makes a particular prediction and allow model developers to debug and improve model performance \citep{ribeiro2016should}.
More generally, explanability holds the promise for enabling appropriate trust in model predictions, detecting discriminatory biases, scientific discovery, and ultimately improving human decision-making.

Meanwhile, the research community has begun to recognize the importance of human perspectives in realizing the promise of explanability.
Explanations of machine learning models serve as a bridge between machines and humans; they are only helpful if they satisfy the need of humans.
In addition to a large body of work on evaluative studies with human subjects \citep{lai2020chicago,lai2019human,green2019principles,green2019disparate,zhang2020effect,poursabzi2018manipulating,wang2021explanations,lage2019evaluation}, \citet{liao2020questioning} interviewed 20 UX and design practitioners working on various AI products to identify gaps between the current algorithmic work and practices for creating explainable AI products.
They developed an explainable AI question bank, representing user needs for explainability as prototypical questions users might ask about the AI (e.g., ``what kind of mistakes is the system likely to make?'' and ``what feature(s) of this instance determine the system's prediction of it?'').
However, it remains an open question {\em how} end-users, such as domain experts or laypeople, are satisfied with current approaches to generating explanations, and---if not---(1) what are the fundamental limitations of the current explanations, and (2) what are desirable approaches to explaining model algorithms in the real world.

Our work fills this gap by retrospectively evaluating the existing approaches to explanability widely adopted by the research community through the lens of real-world decision-makers. In particular, we conducted interviews with domain experts in healthcare and policy-making---including fourteen doctors and twelve policy experts---to understand how they use explanations in their day-to-day work, the pain points they experience with existing explanations, and what would they like to see in the next generation of explanations. The key findings of our qualitative user study are summarized as follows:
\begin{enumerate}
    \item Domain experts are not satisfied with existing explanation paradigms.
    \item Domain experts would prefer an increased interaction with the model about its behavior instead of just seeing one-off explanations such as feature importances or saliency maps.
    \item Domain experts agree that interaction with models and explanations through natural language dialogues would be an advantageous route to more interactive explanations.
    \item Domain experts place a high value on the accuracy/correctness of explanations, yet existing explanations often do not come with an estimate of these metrics.
\end{enumerate} 
Building on the user study, we argue that interactive explanations are a promising avenue for future work for explainability.
We synthesize responses from our interviewees and propose the following principles of interactive explanations: 1) Interact appropriately; 2) Respond appropriately; 3) Properly calibrated responses; 4) Reduce explainability overhead; 5) Consider context.
Next, we point out natural language dialogues could prove to be a powerful tool %
for building an interactive explainability system that satisfies these principles.
 Natural language dialogues can promote accurate and continuous understanding of user queries through rich text interactions and appropriate presentation of explanations, along with understudied problems such as providing confidence and accuracy estimates of explanations. Ultimately, natural language dialogues for explainability could enhance the model's understanding with greater ease than current one-off explanations.
\begin{figure}
    \centering
    \includegraphics[width=\textwidth, trim={0 1.75cm 0 1.75cm},clip]{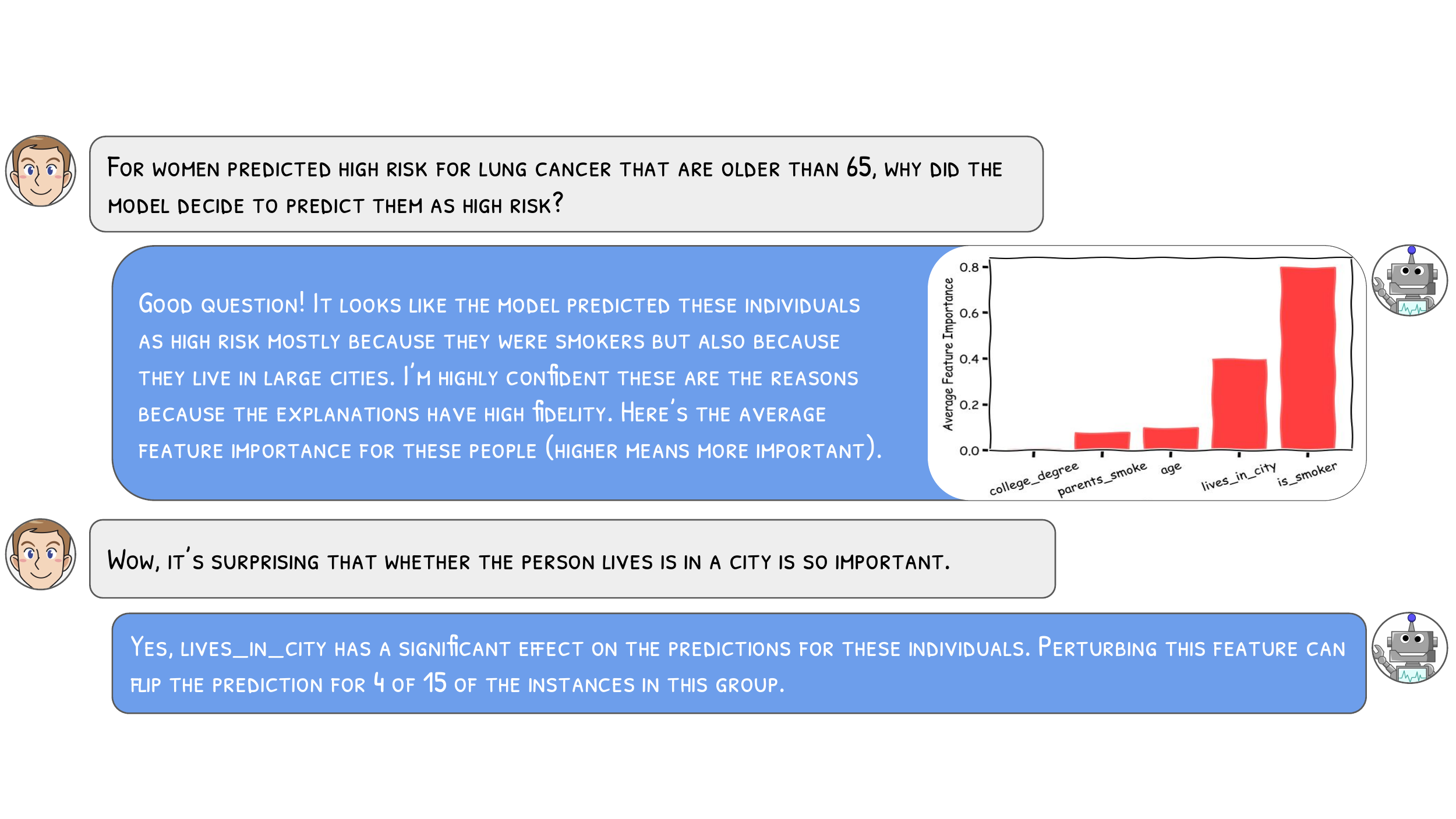}
    \caption{\textbf{A visualization of an example dialogue} users could have with the system for a lung cancer prediction task. The user asks why the model predicted certain individuals who are female and over $65$ as high risk. The model responds with text and visual descriptions and confidence in the explanations. Finally, the user can ask more questions about the explanations. In this case, the user expressed surprise about one of the more important features, and the model runs an additional analysis to demonstrate the effects the feature can have on the predictions for this group.}
    \label{fig:exampledialogue}
\end{figure}
To illustrate these principles, Figure~\ref{fig:exampledialogue} illustrates an instantiation of such natural language dialogues.
A dialogue-based interface provides a novel perspective on interacting with ML models.
We further lay out a possible roadmap towards this exciting vision with four modules: 1) natural language understanding module; 2) explanation algorithm module; 3) response generation module; 4) a GUI interface.
It requires advances in algorithms and human-computer interaction to ensure that these modules enable a satisfying and accurate interaction with models.

Finally, we discuss concrete challenges that we foresee to realize this vision.
We group them into three categories:
1) challenges in natural language understanding given the large set of possible query types, different ways to phrase explainability questions, and complexity due to different applications;
2) challenges in explanation generation that maps intent on correcting explanations and providing confidence estimates for the quality of explanations;
3) challenges in scalability so that the system can respond in real-time.
These challenges serve as an open call for the community to work on them collectively.

In summary, our study identifies the mismatch between current approaches to explanability and the demand of decision-makers in practice, a significant hurdle in adopting complex predictive models in these high-stakes domains.
Our interviews point to natural language dialogues to remedy users' pain points.
We thus propose a research agenda around developing natural language dialogues as the next generation of explanability.
We develop a list of principles, a roadmap towards desirable natural language dialogues, and a set of challenges going forward.
We hope that our work can help the research community realize the promise of explanability and enable effective human-AI interaction with complex machine learning models.

\label{sec:introduction}

\section{Practitioner Perspectives on Rethinking Model Explainability}

In this section, we describe the user study design.
Also, we discuss the results from interviewing $26$ practitioners in the user study and summarize our findings.

\subsection{User Study Design}
Here, we discuss the study that we carried out with practitioners from healthcare and policy to understand how they use explanations in their day-to-day work, the pain points they experience with existing explanations, and what would they like to see in the next generation of explanations. More specifically, we conducted 30-minute long semi-structured interviews with 26 practitioners who regularly employ explainability techniques in their workflow. 14 out of these 26 (53.8\%) practitioners are both medical doctors and researchers who actively use explanation methods to understand ML models that diagnose different kinds of diseases ranging from diabetes to rare cancers. The remaining 12 (46.2\%) of these practitioners are policy researchers who are utilizing explanation methods to understand financial decision making (e.g., loan approvals) models. Furthermore, 18 out of 26 (69.2\%) practitioners are male and the remaining 8 (30.7\%) are female. 16 practitioners (61.5\%) had more than an year of experience working with explainability tools, and the remaining 10 of them (38.5\%) had about 6 months to an year of experience. All the practitioners in our study have used local post hoc explanation methods such as LIME and SHAP in their workflow, and 12 of them (46.2\%) also used various gradient based methods (e.g., GradCAM, Integrated Gradients etc.). 11 participants (42.3\%) also mentioned that they \emph{understand} the technical details of LIME, but none of the participants had any understanding of the inner workings and details of any other explanation methods. 

We began the interview by asking each of the participants about how exactly they leverage model explanations. All the participants said that they look at feature attributions output by post hoc explanation methods for each model prediction of interest, and that they specifically focus on the top 5 to 8 features that are driving the prediction. 21 out of 26 participants mentioned that they also look at the sign (or direction of the contribution) of the feature attribution for certain features of interest---e.g., is salary contributing positively to the loan approval decision? Lastly, 19 out of 26 (73.1\%) participants mentioned that they also compare features w.r.t. their sign, rank, and feature importance values. Our interviews further included, but were not limited to the following questions:
\begin{itemize}
    \item What do you like about model explanations output by state-of-the-art methods?
    \item What do you dislike about model explanations output by state-of-the-art methods?
    \item What other features should explanations have for you to comfortably use them in your day-to-day work? (24 out of 26 participants wanted some form of an interactive dialogue for explanations.)
    \item Would you prefer a one-shot (single) explanation or interactive dialogue style explanations? (We asked this question only if participants did not bring up interactive dialogue on their own; only 2 out of 26 participants did not)
    \item What are the key desiderata you would like to have in interactive dialog style explanations? (We asked this question to the 25 out of 26 participants who wanted interactive dialogue style explanations). 
\end{itemize}

\label{sec:userstudy}

\subsection{Results and Findings}
Our study evaluated users' perceptions about the strengths and weaknesses of current explanations and whether explainability dialogues could help users better understand machine learning models.
This section discusses the respondent's opinions and feedback to this end.
Overall, while respondents were satisfied with many features of current explainability techniques, they pointed to several critical shortcomings with current methods.
Further, respondents expressed a strong desire for interactive explanations and felt that natural language dialogues could serve as an advantageous type of interactive explanations.
Last, interviewees felt that natural language dialogues could create a better explainability experience and identified critical criteria explainability dialogues should satisfy.

\subsubsection{\textbf{The Need for Interactive Explanations}}

During the interviews, respondents indicated several aspects of current explainability techniques they liked.
Respondents most enjoyed getting some understanding of deep learning models (26/26 liked) and understanding which features contribute positively and negatively (21/26 liked). Slightly fewer respondents enjoyed seeing the essential features for predictions (19/26 liked) and comparing the relative importance of features (18/26 liked).
All in all, respondents expressed that current explainability techniques help understand how machine learning models work and how different features affect the model predictions.

While respondents indicated they enjoyed certain features of explainability techniques, they also expressed several unsatisfactory aspects of explanations.
Respondents were most dissatisfied with the lack of additional interaction with explanations after generation.
Respondents answered that they were highly dissatisfied with the fact that conversations with the explanations are not possible (25/26 disliked),  there is no capacity to follow up on explanations (24/26 disliked) interactively, nor ask custom questions (23/26 disliked).
One respondent stated, ``It is extremely frustrating to just look at one explanation [per prediction] and not be able to follow up on it!''
Another indicated, ``I should be able to ask custom questions [to the explanation] and get answers.''
Respondents also disliked that they could not understand the accuracy of explanations (24/26 disliked).
One of the interviewees described, ``I don't know anything about how correct the explanation is! How do you expect me to use it meaningfully? I constantly struggle with worrying about using an incorrect explanation and missing out on not using a correct explanation that is giving me more insights.''
Slightly fewer respondents indicated they disliked the limited capacity of explanations to generate subgroup-level explanations (21/26 disliked).
One respondent questioned, ``Why is all explanation work focused on local explanations? I would like to see at least subgroup level explanations. I think there is one algorithm (MUSE?) but need a lot more work.''
Overall, respondents expressed evident dissatisfaction with the one-off nature of explanations.
Respondents felt that, in almost all cases, they had further follow-up questions for explanations to do with the explanation's accuracy or additional tasks they would like the explanation to solve.
The interviewees felt that the lack of interactivity with explanations is a significant shortcoming of existing techniques.

When we asked respondents what could improve explanations, respondents discussed several potential improvements.
Overall, respondents expressed the strongest desire for explainability through fully-fledged conversations with ML models (25/26 said this was important).
Multiple respondents voiced support for conversational explanations.
One stated, ``I can see myself using explainable tools a ton more if only it were like a free-flowing dialogue. Oh I can't wait for that day.''
Another said, ``dialogue-based explanations will totally revolutionize how medical science uses ML. Wow, I am excited just thinking about the possibility.''
Respondents also indicated the inclusion of reliable accuracy metrics for explanations as a critical place of improvement (24/26 said this was important).
Slightly fewer respondents indicated that custom questions are vital for improving explanations (22/26 said this was important) and improving subgroup level explanations (24/26 said this was important).
While the interviewees indicated several places to improve explanations (such as accuracy metrics), respondents most heavily fixated on the potential of having conversations with ML models to support explainability.
Considering that respondents indicated a strong desire for dialogue-based explanation systems, we next discussed key desiderata for an explainability dialogue system.

\subsection{\textbf{Explainability Dialogue Desiderata from Interviewees}}

 Respondents felt that fully-fledged conversations in natural language with explanations would help them better understand ML models.
In addition, they felt that a explainability dialogue system could greatly help their explainability workflows.
 Further, respondents had numerous ideas about the system's capabilities, different ways the system could augment their explainability workflows, and what they hoped to get out of such a system.

 Critically, many respondents envisioned explainability dialogues happening much like conversations with colleagues where the goal is to understand ``why'' another practitioner made a particular decision or choice (e.g., medical diagnosis, financial risk assessment).
 In this sense, they imagined treating models like colleagues and using explainability dialogues to facilitate natural interactions between models and people.
Respondents viewed such natural language conversations as more intuitive for understanding model decisions than writing and debugging cumbersome code to generate explanations.
Further, they imagined explainability dialogues giving more context to the explanations, such as assessments of accuracy,  descriptions of how to interpret the explanations, and uncertainty, much like people do in everyday conversations~\cite{paek2000conversationactionuncertainty}.
Finally, they viewed conversations happening in a context-dependent manner, where they could easily follow up on previous queries for additional clarification or further lines of questioning.
 We summarize the key desiderata agreed on by the respondents below, in order of requirements most respondents agreed was important, where (N/26) indicates the number that agreed:
\begin{itemize}
    \item (24/26) The dialogue should eliminate the need to learn and write the commands for generating explanations.
    \item (24/26) The system should describe the accuracy of the explanation in the dialogues.
    \item (23/26) The system should preserve context and enable follow-up questions.
    \item (21/26) The responses should be provided in real-time.
    \item (17/26) The dialogue system should decide which explanations to run. Users should not have to ask for a specific explainability algorithm.
\end{itemize}
These desiderata capture key elements in respondents' ultimate goals of engaging in conversations with machine learning models.
For instance, respondents were excited about explainability dialogues involving natural, everyday questions to machine learning models such as, "why did you make this decision?" and therefore agreed such systems should eliminate the need to write code, take the conversation context into account, and happen in real-time.
Overall, respondents felt that natural language explainability dialogues would greatly improve their experiences using explanations and had clear ideas about how such a system should behave.

\section{Principles of Interactive Explanations via Natural Language Dialogue}

Leveraging our findings from the interviews, we outline a set of principles interactive explanations should follow.
Based on these principles, we suggest natural-language explainability dialogues as a promising solution to enabling interactive explanations.
As a starting place for further research in this direction, we suggest a concrete design for an explainability dialogue system.

\subsection{Principles of Interactive Explanations}

Given the need for explanations to enable better interactions with models through custom queries, additional follow-up questions, and proper contextualization, there are exciting research opportunities for developing interactive explanations for machine learning models.  
Interactive explanations should enable rich and continuous interactions with models that enable users to understand how their models work.
Further, users should engage with interactive explanations in ways that are not frustrating, require minimal or no coding overhead, and facilitate improved model understanding as they use the system.
Also, interactive explanations should improve users' ability to correctly utilize explainability techniques to understand how trustworthy models are and interpret the results of explanations.
Finally, interactive explanations should give users a properly calibrated sense of trust in their models, encouraging trust when continued interactions reveal their models are right for the right reasons and reducing trust when this is not the case.

Considering the interviews and our perceptions about what interactive explanations should look like, we now propose five principles for designing an interactive explanation system as a starting place for research in this direction.
The principles are as follows:
\begin{itemize}
\item \textbf{Principle 1 (Interact Appropriately).} The system should understand continuous requests for explanations and be able to efficiently map these to appropriate explanations to run.

\item \textbf{Principle 2 (Respond Appropriately).} The system should respond with informative, properly contextualized, and satisfying explanations for why the model made specific decisions.
\item \textbf{Principle 3 (Properly Calibrated Responses).} The system should provide reliable notions of confidence along with explanations.
\item \textbf{Principle 4 (Reduce Explainability Overhead).} The system should reduce or eliminate the need for users to write code to explain machine learning models. The system should strictly make understanding machine learning models easier for users.
\item \textbf{Principle 5 (Consider Context).} The system should condition its understanding of inputs on the previous interactions, including prior inputs, responses, and data sets among potentially other artifacts generated in the interaction.
\end{itemize}

Principle 1 is critical for the system to comprehend users' inputs and map them to appropriate explanation outcomes. The system must understand a wide range of queries and how to act on each of them appropriately.
Principle 2 ensures the users can understand the explanations provided by the system.
This principle is essential because users should easily comprehend the system outputs and retain the natural flow of the interactions.
Principle 3 is important because current explanations often do not adequately contextual responses. Ideally, the system should provide confidence or accuracy associated with the explanation so that users will know where to trust the explanations.
Principle 4 is paramount because adding an interaction layer on top of explanations inherently creates further technical complexity.
Consequently, this increases the risk of errors by misunderstanding user inputs or running the wrong explanations.
The benefits of enabling interactive explanations should outweigh any potential issues and complexities of implementing an interactive explanation system.
Principle 5 is important for ensuring the system engages in natural interactions with users.
The interactions should build on themselves, establishing different threads used to condition future responses where appropriate.

Overall, Principle 1 and 2 speak to the quality of single-round explanations; 
Principle 3 highlights new capabilities that prior works have overlooked;
Principle 4 weighs the benefit of natural language dialogues against the risks;
Principle 5 promotes multi-round conversations.
We recommend that designers of explainability dialogue systems use these principles as a starting place when deciding whether and how to implement such a system.

\subsection{Roadmap towards an Explainability Dialogue System}

Considering the need for an interactive explainability system and favorable opinions of the interview respondents, we suggest natural language dialogues as an appropriate way to accomplish interactive explanations.
Natural language dialogues satisfy principles \textbf{(1-5)}, making them an ideal choice for such a system.
For instance, dialogues can handle a diverse set of requests, can offer dynamic responses, and are inherently contextual (Principles 1, 2, 5).
Further, it is possible to include extensive context for explanations generated by the system in natural language (Principle 3). 
Last, by enabling machine learning models to be questioned in natural language, like another colleague, explainability dialogues will make it straightforward for anyone to understand ML models (Principle 4).

Moreover, a dialogue-based explanation interface provides a novel perspective on interacting with ML models. Rather than treating a model as an object that only returns decisions for inputs, we can think of models as entities that anyone can interact with in natural language.
This allows models to be ``accountable'' in the sense that anyone can query them for a justification behind decisions.
As we see in the interviews, domain experts that use machine learning models have a strong desire to treat models as another colleague, i.e., an entity that can be asked, in natural language terms, for a decision and justification. 
In this way, domain experts wish to decide whether or not to trust machine learning models in a manner that is more accessible and natural than using current explainability techniques out of the box. 

Implementing an explainable dialogue system that satisfies principles \textbf{(1-5)} involves designing several non-trivial technical components that touch on a wide array of technologies.
Likely it will require building both supervised and generative natural language processing models, a wide variety of different explanations, and novel improvements to explanations.
In addition, ensuring that the system is available and can rapidly serve responses will require advances in efficiency (e.g., distributed systems).
Finally, ensuring that users can interact with the dialogue system satisfactorily  will require further HCI research and user studyies.

One way to design an explainability dialogue system is by separating the system into four modules: 
\begin{enumerate}[label=\bfseries(\roman*)]
\item A natural language understanding module that understands the user input and determines what explanation(s) to generate.
\item An explanation module that runs the explanations.
\item A response generation module that serves a response to the user. 
\item A GUI interface for the system.
\end{enumerate}
For this system design, we will consider designing them independently, though it could be possible to design modules \textbf{(i-iii)} in an end-to-end manner.

For module \textbf{(i)}, dialogue system designers could train two large language models (LLMs) \cite{devlin-etal-2019-bert, 2020t5}. The first could predict what filtering operations to run (i.e., get instance \texttt{id=0}) in a semantic parsing style ~\cite{he:20a}. 
This step would determine which instances to explain.
A second model could predict what explanation to run out of a finite set of possible explanations.
To consider the context of the conversation, designers could condition these two models on the previous text in the conversation of fixed window size.
Finally, training these models would require generating two separate datasets. 
The first model would require a standard text to SQL dataset, like Spider or WikiSQL~\cite{Yu18c, zhongSeq2SQL2017}. 
The dataset for the second model would need potential input queries and appropriate filtering \& explanation responses.
Also, it would be necessary to augment both of these datasets with examples of context from conversations.

With the explanations to generate and filtering operations in hand, module \textbf{(ii)} runs the explanations.
One complication with this step is that if users request many explanations, generating explanations could be a bottleneck in the system.
It will likely be necessary to batch out running explanations across a set of machines or servers to ensure the system rapidly generates them.

For generating responses in module \textbf{(iii)}, designers could train a generative LLM in a fact-aware manner to return rich text outputs from the system that also include factually correct explanation responses~\cite{logan-etal-2019-baracks}.
Because visualizations are vital components of explanations, it will also be likely that system designers will include the visualizations from the explanations~\cite{lundberg2017unified}.

Finally, to facilitate users providing text input and serving responses, it will be necessary to design a graphical user interface (GUI).
It could be possible for ML-friendly UI packages such as Gradio~\cite{abid2019gradio} to create such an interface.
We envision what a conversation in such an explainability dialogue system could look like in Figure~\ref{fig:exampledialogue}.
The user provides the system with a high-level question about why the model predicts women older than sixty-five as high-risk for developing lung cancer.
The system understands the user's request to generate feature importance explanation across this demographic.
The system performs the filter operations necessary to get these instances and runs the feature importance explanations.
Last, it generates a helpful summary of the operations and gives them to the user.
Finally, the user indicates surprise surrounding one of the features, \texttt{lives\_in\_city} being important.
The system understands the user's hesitation and provides further validation for the claim to the user.
Overall, the dialogue system correctly handles the user's questions, provides valuable responses, and understands sufficient context in the conversation to handle further follow-ups.

\label{sec:desiderata}

\section{Natural Language Dialogues for Explainability: Risks and Research Opportunities}

While there are concrete approaches to designing natural language dialogues for explainability, there are numerous challenges in implementing such a system that motivate several research opportunities.
We divide our discussion about the risks and opportunities into four parts.
First, we focus on the natural language processing aspect of the system, including the language understanding and generation components.
Second, we examine the explainability aspect of the system.
Next, we evaluate the interface and UI component of the dialogue. 
Last, we discuss the scalability and real-time response needs of the explainability dialogue system.

\subsection{Language Understanding Considerations}

In this subsection, we discuss challenges due to understanding natural language in explainability dialogues.

\subsubsection{Understanding Language in Explainability Dialogues} A fundamental difficulty in developing an explainability dialogue system is that model designers ask many complex questions when interacting with machine learning models.
Developing a system capable of understanding various user questions is difficult given the broad domain of possible queries, their complexity, and the different ways users might structure them.
For instance, why does a model make predictions across the entire domain, for groups of instances, or individual instances? 
Does the model learn intuitive rules for prediction or something more complicated? 
What parts of the model are most important for predictions? 
What data is most useful for learning? 
In what ways do you have to change instances to get alternative predictions? 
The wide variety of questions users of an explainability dialogue system will ask dramatically adds to the complexity of developing such a system.

There are many possible ways to structure the language in the explainability dialogue in addition to application-specific terminology.
These variations can come in the form of different semantics.
For instance, a user might ask ``what are the most important features for this prediction'' or ``what inputs did the model rely on when making this classification.''
In both cases, the explainability dialogue system must understand that the user requests a feature importance explanation.
Users can also ask questions with different levels of specificity.
A user concerned with running a particular explanation might ask, ``please provide LIME feature importance explanations for data points with id's 10-15.''
Users will also likely ask high-level questions such as ``what is the reason for the predictions for my data.''
In both these cases, the explainability dialogue system must correctly understand the user questions and map them to an appropriate outcome.

Finally, explainability dialogues are complicated by the application-specific nature of explaining machine learning models.
In a medical application, users will ask questions about different features and outcomes than in a finance application.
For instance, users of such a system applied to a medical application will ask whether medical history had to do with specific predictions by the model. In contrast, financial applications will likely ask about income or employment history.
Designing a satisfactory explainability dialogue system involves developing the system to handle a wide variety of application-specific language.

Given the large set of possible query types, complexity due to different applications, and different ways to phrase explainability questions, developing natural language processing systems that can understand explainability dialogues is technically challenging.
Ideally, such a system should be capable of understanding a wide variety of explainability dialogue and quickly adapting to new, domain-specific terminology.
LLM's may be unequipped out-of-the-box to handle explainability dialogues, due to deficiencies in their training data for such tasks for instance.
Consequently, additional work may be necessary to adapt LLM's to explainability dialogue.

\subsubsection{Allowing Rich Multi-Turn Dialogue}

Another technical challenge with explainability dialogue systems is developing multi-turn conversational systems---those that use the context of the conversation to inform future responses.
Though many \textit{single-turn} conversational AI systems exist that do not leverage the previous context (e.g., Siri), ideally an explainability dialogue system should be capable of leveraging the entire conversation to inform future responses.
Leveraging context to inform the dialogue responses enables richer and more natural conversations with the explainability dialogue system.
For example, it is highly desirable to compare and contrast with previous explanations within a conversation.
Users might want to ask, ``Does the model rely on similar features for this instance as it did with the previous explanation?'' or ''if I changed this data point, how would the explanation change?''
In both these situations, it is necessary to maintain the state of the conversation to inform the future response.

Enabling multi-turn conversations presents a number of significant technical hurdles.
For instance, it could be possible to condition LLM's on the conversation.
However, LLM's typically have a fixed window size for text, making it difficult to include the entire conversation.
Consequently, this motivates difficult technical decisions, such as whether to include a finite window size of the conversation or select parts of the conversation to condition on in some intelligent manner.
Of course, selecting which parts of the conversation to condition on increases the likelihood critical parts of the conversation will not be included by error and consequently harm the user-experience.

\subsubsection{Responding Appropriately in Dialogue}

An additional technical challenge is determining how to appropriately generate responses to user inputs in the explainability dialogue.
Ideally, responses provided by the explainability dialogue system will be highly flexible, dynamically generated for any given user response, sensitive to the tone of the conversation and the expected effects of the response, seek additional information from the user when necessary, and present explanations in both a visually and semantically satisfying way.
Nevertheless, generating such responses in dialogue is still an open problem using current state-of-the-art language models, such as LLM's.
Though LLM's generate realistic text outputs, allowing free-form natural language responses in dialogue is unpredictable, often forgets context, and can lead to incorrect or, in the worst case, offensive responses~\cite{zaib2020shortsurvey, chen2017recentadvancesdialogue, oluwatobi-mueller-2020-dlgnet, zhang2019dialogpt}.
In settings where it is not acceptable to have unpredictable and potentially harmful responses, it may not be appropriate to use dynamically generated responses by LLM's.
Consequently, it could be necessary to use scripted responses, in order to enforce the trustworthiness of the explainability dialogue system.

Still, using scripted responses presents a number of significant considerations and shortcomings.
As we have discussed, there are numerous different types of explanations and creating acceptable responses for all of these explanations is a challenging and laborious task. 
For example, explanations are often presented visually (e.g., the SHAP feature importance plots \cite{lundberg2017unified}).
Determining an acceptable way to present visual explanation responses in conversations along with sufficient textual explanations so that the user will understand the explanations requires careful consideration.
Further, scripted dialogues may lack the flexibility to enable meaningful interactions in multi-turn conversations.
Last, there are opportunities for a middle ground between completely dynamic explainability dialogue systems and fully scripted ones, in ways that can satisfy both reliability and flexibility goals.
Overall, designers of explainability dialogue systems must carefully weight the trade-offs between dynamically generating responses and hard-coding them.

\subsection{Explainability Considerations}

This subsection discusses explainability considerations in the dialogue system. 
First, we focus on what explanations to use in such a dialogue system.
Next, we look at issues with current explanations and how these may cause complications in a dialogue.
Finally, we discuss how current explanations are unsatisfactory for dialogue and opportunities for improving explanations.

\subsubsection{Mapping Intent to Correct Explanations}

Once the system understands user intent within the dialogue, a critical technical consideration is mapping the intent to an appropriate set of explanations to generate for the conversation.
Foremost, mapping intent to explanations is complicated because multiple different explanations could provide a satisfactory response to a user request.
For example, in conversations where users request the important features for prediction, the dialogue system could provide any feature importance explanation. 
Considering that there are numerous different types of feature importance explanations (LIME~\cite{ribeiro2016should}, BayesLIME~\cite{reliable:neurips21}, SHAP~\cite{lundberg2017unified}, Smooth Grad~\cite{smilkov2017smoothgrad}, etc.) each with different trade-offs, it is difficult to decide what explanations to provide.
For instance, LIME explanations are often relatively quicker to generate than SHAP explanations.
However, SHAP provides a more robust implementation and better support.
These trade-offs must be considered when choosing between what (or what set of) feature importance explanations to provide in the explainability dialogue.

Ultimately it is the responsibility of the explainability dialogue system designer to make decisions regarding which explanations to provide in the dialogue.
This decision is not straightforward because explanations within the same categories (i.e., feature importance, counterfactual explanations, global decision rules) make differing assumptions and often provide different results.
The designer will need to consider the applications of the system, technical implementations of the explanations, and fundamental trade-offs between the different types of explanations to make an informed decision around what explanations to provide.

\subsubsection{Shortcomings of Current Explanations}

Though explanations are beneficial for informing users why a model makes decisions, current techniques suffer from several shortcomings.
If used in explainability dialogue, these shortcomings will likely translate into the system, potentially causing issues by providing misleading explanations or causing significant system slowdowns.
These shortcomings include that explanations are unstable \cite{ghorbani2019interpretation, SlackHilgard2020FoolingLIMESHAP, dombrowski2019explanations, adebayo2018sanity, AlvarezMelis2018OnTR}.
Meaning, slight perturbations to instances can result in significant changes to the explanation.
In addition, it is difficult to set the hyperparameter values of explanations (the number of perturbations in LIME \cite{ribeiro2016should} \& SHAP \cite{lundberg2017unified} or the kernel width in LIME).
These hyperparameter choices can have significant effects on the explanation.
Further, they are inconsistent~\cite{lee2019developing}---rerunning explanations can lead to different results.
In addition, there are few metrics to determine the quality of the explanations, making it challenging to decide when to disregard generated explanations.
Finally, specific explanations (e.g., LIME \& SHAP) are incredibly time intensive to generate because they rely on repeatedly querying the model~\cite{chen2018lshapley}.
Designers of explainability dialogue systems must consider the different limitations of current explanation techniques and evaluate how they will affect the system's performance.

\subsubsection{Opportunities For Improving Explanations}

There are several opportunities to improve explanations to ensure they are satisfactory for an explainability dialogue.
Designers of explainability dialogues should focus on developing consistent explanations and ensuring that the same query repeated does not return different results.   
In addition, designers of such systems should consider how to generate confidence or accuracy metrics for explanations.
There is some work in this direction in the form of Bayesian Local Explanations by~\citet{reliable:neurips21} that generate local explanations along with associated confidence values.
However, extensive further work is needed to determine confidence metrics for additional types of explanations beyond local explanations.
Further, designers should develop techniques to set the explanation hyperparameters without requiring intervention from users.
This direction also motivates developing hyperparameter-free explanations.
Systems designers could more easily incorporate hyperparameter-free explanations in dialogue systems to eliminate the need to set hyperparameters throughout the dialog.
Finally, researchers should consider developing explanations that are more accurate in the first place, making them less error prone at the start.
Overall, there is considerable opportunity for developing novel explainability techniques that make explanations more compatible with dialogues.

\subsection{Interface Considerations}

An interface to the explainability dialogue system should facilitate users providing text inputs and rapidly receiving responses from the back-end system in a straightforward manner to understand.
Depending on the application, it could be possible to have either a text-based interface or a spoken-word dialogue system.
For operational applications where users may be unable to take the time to type out sentences, a spoken-word system may be appropriate.
However, adding a step to convert spoken words into text adds further complexity and room for error.
An explainability dialogue system in the text will likely suffice in many scientific, engineering, or corporate applications.
Also, because explanations often involve visual components, it should be possible for the interface to display images, plots, and tables.
Finally, the designers should build an interface that is easily accessible for users, such as a web application compatible with desktop computers, tablets, and cell phones.
Altogether, the interface for the dialogue system should be easy to use, both in terms of its accessibility and presentation.

\subsection{Scalability Considerations}

Ideally, an explainabiltiy dialogue system should yield explanations without lag time from when the user provides input.
However, there are several technical considerations that influence the response time and scalability of such a system.
These considerations can be divided between the NLP and explainability components of the dialogue system.
First, we discuss NLP scalability considerations. After, we cover explainability concerns.

Designers will likely use LLM's in the explainability dialogue system to understand complex natural language inputs and yield rich text responses. Because LLM's have many parameters, GPU acceleration is necessary. GPUs can increase the burden of running and scaling the system due to their cost, lack of availability, and difficulty to maintain. Further, querying parameter-intensive LLM's can be time-consuming, even with GPU acceleration, slowing down the system's response time. Finally, it may be necessary to use state-of-the-art LLMs to achieve acceptable performance at explainability dialogue. System designers cannot easily run these models on available hardware because of their high parameter counts, and designers may have to use costly APIs provided by private companies (e.g., GPT3 \cite{brown2020llmfewshot}).
Designers will need to carefully weigh NLP modeling choices with hardware availability when designing dialogue systems so that they are not a bottleneck in scaling and running explainabilty dialogue systems.

In addition, there are potential scalability issues with the explanations used in the system.
Various explanations have prolonged run times, which could adversely affect the system's response time.
For example, model-agnostic feature importance explanations such as LIME~\cite{ribeiro2016should} and SHAP~\cite{lundberg2017unified} are notoriously slow due to their repeated querying of the black-box model \cite{reliable:neurips21}.
If used to generate feature importance explanations in an explainability dialogue, such explanations could lead to long response times between when the user provides input and the system responds, especially if the black-box model is complex or the users ask for many explanations.
Though explanations have slow runtimes, there are potential technical solutions.
For instance, the system could generate and cache explanations in the background during the conversation.
System designers could further accelerate explanation generation through parallelism.
If users request explanations that are cached, the system can use them in the dialogue immediately.
Of course, introducing caching and parallelism further complicate the system's complexity and resource requirements.
Overall, system designers will need to consider how to best handle slow explanation run time to ensure real-time explainability dialogue, given the system's expected needs and resource constraints.

\label{sec:technicalconsiderations}

\section{Related Work}
\paragraph{Interpretability Techniques} Work on machine learning explanations includes two main directions: inherently interpretable models and post hoc explanations.
Researchers have proposed models that are \textit{inherently interpretable}---i.e., those models that are interpretable by design.
Inherently interpretable methods include decision lists and sets~\cite{lakkaraju2016interpretable, angelino2017learning}, additive models \cite{ustun2013supersparse, lou2013accurate}, and prototype based models~\cite{kim2015bayesian, chen2019thislookslikethat, Li2018DeepLF}.
However, inherently interpretable models constrain model designers to specific models that may lack sufficient expressiveness for complex tasks.
Consequently, there has been considerable recent interest in post hoc explanations.

On the other hand, post hoc explanations provide explanations for machine learning models that have already been trained, allowing greater flexibility in the modeling process.
There are several different types of post hoc explanations. These include \textit{model agnostic methods} that do not rely on access to model internals such as LIME~\cite{ribeiro2016should} \& SHAP~\cite{lundberg2017unified, pmlr-v130-covert21a}, BayesLIME \& BayesSHAP~\cite{reliable:neurips21}, partial dependency plots~\cite{friedman2001greedy}, and permutation feature importance~\cite{breiman2001random}. Also, there are post hoc explanations, which assume access to model internals (e.g., gradient access)~\cite{smilkov2017smoothgrad, selvaraju2017grad, sundararajan2017axiomatic, simonyan2013deep}. 
There are also global post hoc explanations.
These methods summarize the model's decision logic across part of or the entire domain into interpretable rules or decision trees~\cite{bastani2017interpretability, lakkaraju2016interpretable, kim2017interpretability}.
Last, there has been considerable recent interest in \textit{counterfactual explanations} that describe changes to instance which will result in different model predictions~\cite{wachter2017counterfactual, ustun2019recourse, poyiadzi2020face, proto20, barocas2020hiddenassumptions, KarKugSchVal20, KarSchVal20}.
Model providers can leverage counterfactual explanations to provide individuals adversely affected by model decisions with recourse.
Though there are numerous works on developing explanations, few works have considered interactive explanations in the form of dialogue.

\paragraph{HCI Analyses of Explanations}

Within the human-computer interaction (HCI) literature, several studies examine how model designers design, build, and correct ML models, finding that interpretability and interactivity are critical for iterating on ML models~\cite{adbdul2018, dourish2016algorithmsandtheirothers, hohman2019gamut, bellotti2001intel}.
\citet{fails2003interactiveml} first propose the term ``interactive machine learning'' for systems where users train ML models and correct their predictions.
Numerous interactive machine learning exist, including general purpose systems like Orange~\cite{orange2004} and application-specific platforms such as Abstrackr for citation review~\cite{wallacedeploying2012}.
Additional works study to what extent ML explainability techniques help data scientists and find data scientists trust explanations too much or do not use them in the correct way~\cite{kaur2020interpreting}. Further works evaluate the interpretability of certain classes of machine learning models and determine that humans have an easier time simulating models with fewer features, fewer parameters, and access to the model's internals. 
However, people still struggle to decide when to trust the model predictions, even if they can simulate it~\cite{Slack2019AssessingTL, PoursabziSangdeh2021ManipulatingAM}. 
Furthermore, a growing set of literature adopts application-based evaluation and examines the impact of explanations on human-AI decision making \citep{lai2020chicago,lai2019human,green2019principles,green2019disparate,zhang2020effect,poursabzi2018manipulating,wang2021explanations,lage2019evaluation}.
In particular, \citet{liu2021understanding} studies the effect of interactive explanations that allow users to change the input and observe the differences in the output.
To the best of our knowledge, there is little work in allowing interaction through natural language dialogue for explainability.

\paragraph{ML Analyses of Explanations}
There have been several critical analyses of explanations from within the ML literature.
Foremost,~\citet{rudin2019stop} makes the case that post hoc explanations are inherently unfaithful to the model. 
Instead, model designers should build inherently interpretable models.
Other works examine the robustness of explanations from both theoretical and empirical perspectives~\cite{garreau2020looking, levine2019certifiably, pmlr-v119-chalasani20a, ghorbani2019interpretation, pmlr-v139-agarwal21c, zafar-etal-2021-lack, lai+cai+tan:19}.
\citet{AlvarezMelis2018OnTR} show that explanations are unstable and small perturbations can lead to drastically different explanations.
\citet{zafar-etal-2021-lack} demonstrates models only differing in their initialization can have distinct and sometimes contradictory explanations.
Further works demonstrate that malicious adversaries can manipulate explanations, demonstrating their unreliability~\cite{fairwashingoffmanifold, facade:femnlp20, Dimanov2020YouST}.
For instance,~\citet{SlackHilgard2020FoolingLIMESHAP} show adversaries can design models that how arbitrary LIME~\cite{ribeiro2016should} \& SHAP~\cite{lundberg2017unified} explanations.

\paragraph{Dialogue Systems}

There has been considerable work in developing dialogue systems in the past decades~\cite{Arora2013DialogueSA, WANG2016303, chen2017recentadvancesdialogue, gao-etal-2018-neural-approaches}.
Currently, state-of-the-art approaches rely on deep learning systems to understand user inputs and for generating responses~\cite{Ni2021RecentAI}.
Work in dialogue systems can be divided into task-oriented dialogues and open-domain dialogues.
Task oriented dialogue systems address a particular problem, such as scheduling an appointment or making a reservations.
Open-domain dialogues do not solve a particular task and instead attempt to ``chit-chat'' with users.
We focus on task-oriented dialogues because they are most relevant to an explainability dialogue system.
There are several different approaches to task oriented dialogue systems~\cite{bordes2017endtoendialog, williams-etal-2013-dialog, el-asri-etal-2017-frames, zhu-etal-2020-crosswoz, rastogi2019towards, byrne-etal-2019-taskmaster, peskov-etal-2019-multi, yu-etal-2019-cosql}.
Certain systems use separate modules for language understanding, state tracking, planning, and response generation to accomplish task oriented dialogues~\cite{henderson-etal-2013-deep, Henderson2015MachineLF, kim2018twostep, singla-etal-2020-towards}. 
Other systems use fully end-to-end approaches for task oriented dialogue~\cite{le-etal-2020-uniconv, wang-etal-2019-incremental, ham-etal-2020-end}.
There are several trade-offs between modular and end-to-end approaches.
With modular approaches, it is difficult to propagate errors in the system response to all the system modules \cite{Ni2021RecentAI, le-etal-2020-uniconv}.
Instead, designers must carefully design each of the modules.
However, end-to-end approaches require sufficient data in order to have high response quality, making it difficult to use them in data-scarce settings~\cite{balakrishnan-etal-2019-constrained, kale-rastogi-2020-template}.
Further, when errors due occur, it is often more difficult to debug end-to-end dialogue systems.
Designers of explainability dialogues will need to carefully consider the availability of data for such systems when deciding which approaches to pursue.

\label{sec:relatedwork}

\section{Discussion \& Conclusions}

Natural language dialogues are a promising approach for facilitating interactive explanations of machine learning models.
Language is an ideal medium to engage with machine learning models because of its flexibility and accessibility, enabling anyone to understand ML models.
Consequently, explainability dialogue systems could enable rich interactions with models through complex, high-level queries and in-depth, contextualized responses consisting of both text and visual artifacts.
While explainability dialogues are appealing for domain experts or users with limited machine learning knowledge, these systems are still valuable to experts because of their ease of use and capacity to rapidly facilitate understanding models in many ways.
As a result, explainability dialogues could serve a vital role in enabling model understanding for any stakeholder in an ML model.

While there are numerous advantages of natural language as a solution to interactive explanations, it is important to understand its limitations in various application scenarios.
First, different domains may need specific interactions to appropriately engage with the models and data.
For example, users will likely ask questions to an explainability dialogue system for images that reference specific parts of the image such as, ``is the nose of the dog an important feature'' or ``is the model using the upper right-hand side of the image?''
It is challenging to understand what parts of the image the user refers to with natural language.
However, it could be possible to develop multimodal models capable of handling such requests.
To better suit these domains, it could also be possible to incorporate other types of interactions, such as clicking on parts of an image, into the dialogue system interface.

Similarly, some system responses may be difficult to communicate in natural language.
For example, it is pretty challenging to communicate uncertainty in natural language, making it difficult for a dialogue system to explain uncertainty.
One solution is to use plots or other visuals to present difficult to communicate concepts.
Designers can easily include these in an explainability dialogue, like in Figure~\ref{fig:exampledialogue}.

In addition, domain experts engaging in multiple lines of inquiry with a machine learning model may wish to pick up early threads, drop recent ones, or switch between many different threads in their inquiry.
However, a natural language dialogue is a linear interaction, where the system and users take turns responding.
This structure makes it challenging to have multiple threads in the conversation.
However, it is possible to include a multi-threaded conversation feature, much like \textit{Threads} in the Slack messaging application or \textit{Replies} in iMessage.

Finally, language can often be vague.
Questions are often under-specified, forcing respondents to assume what the questioner wants.
Often, it is necessary to ask follow-up questions.
Also, different people may have completely different intentions when they ask the same thing.
This ambiguity makes it difficult to understand what the questioner wants even though the system fully comprehends the text provided.
A natural language dialogue system will likely need to follow up on underspecified questions.
When users cannot specify what they want in words, it may be necessary for the system to provide access to a command line or coding interface to request what they want.

Overall, explainability dialogue systems deserve considerable attention from the research community.
Such systems could revolutionize how domain experts and users interface with machine learning models, facilitating the safer use of machine learning models.
We encourage the research community to develop explainability dialogue systems and enable more accessible model understanding.

\label{sec:discuss_conclude}

\bibliography{bibliography}

\end{document}